\documentclass[article]{IEEEtran}
\IEEEoverridecommandlockouts

\usepackage{cite}
\usepackage{amsmath,amssymb,amsfonts}
\usepackage{algorithmic}
\usepackage{graphicx}
\usepackage{textcomp}
\usepackage{xcolor}
\usepackage{hyperref}
\usepackage{caption}
\usepackage{subcaption}
\usepackage{multirow}
\usepackage{float}
\usepackage{placeins}
\usepackage{lipsum}
\usepackage{graphicx}
\usepackage{subcaption}
\usepackage{mwe}
\usepackage{diagbox}
\restylefloat{table}
\begin{document}

\title{RGait-NET: An Effective Network for Recovering Missing Information from Occluded Gait Cycles}

\author{\IEEEauthorblockN{Ayush~Agarwal\textsuperscript{1}}
\IEEEauthorblockA{Motilal Nehru National Institute of Technology, Allahabad, 
\emph{ayushagarwal@mnnit.ac.in}}\\
\and
\IEEEauthorblockN{Dhritimaan~Das\textsuperscript{1}, }
\IEEEauthorblockA{Indian Institute of Technology (BHU), Varanasi, 
\emph{dhritimaand.cd.eee17@itbhu.ac.in}}\\
\and
\IEEEauthorblockN{Pratik~Chattopadhyay\textsuperscript{2}, }
\IEEEauthorblockA{Indian Institute of Technology (BHU), Varanasi, 
\emph{pratik.cse@itbhu.ac.in}}\\
\and
\IEEEauthorblockN{Lipo~Wang\textsuperscript{3}}
\IEEEauthorblockA{Nanyang Technological University, 
Singapore, 
\emph{ELPWang@ntu.edu.sg}}}

\maketitle

\begin{abstract}
\normalfont Gait of a person refers to his/her walking pattern, and according to medical studies gait of every individual is unique. Over the past decade, several computer vision-based gait recognition approaches have been proposed in which walking information corresponding to a complete gait cycle has been used to construct gait features for person identification. %generally revolve around the idea to extract unique features from the walking sequence for each subject and use them for classification. Since the data for gait recognition is in the from of video sequences from which frames are extracted, 
Majority of these methods compute gait features with the assumption that a complete gait cycle is made available always. However, in most public places occlusion is an inevitable occurrence, and %so in real life occlusion in data in form of videos is inevitable due to crowding in public places like airports, railway stations etc.. Also o
due to this, only a fraction of a gait cycle gets captured by the monitoring camera. Unavailability of complete gait cycle information drastically affects the accuracy of the extracted features, and till date, only a few occlusion handling strategies have been proposed in the gait recognition literature. However, none of these performs reliably and robustly in the presence of a single cycle with incomplete information due to which practical application of computer vision-based gait recognition is quite limited. In this work, we improve the state-of-the-art by developing novel learning-based algorithms to identify the occluded frames in a given gait sequence, and then predict the missing frames to reconstruct the gait cycle. Specifically, occlusion detection has been carried out by employing a VGG-16 model, and a Long-Short Term Memory network is trained to optimize a multi-objective loss function based on dice coefficient and cross-entropy loss to reconstruct the occluded frames. The effectiveness of the proposed occlusion reconstruction algorithm has been evaluated based on the Gait Energy Image (GEI) feature on the reconstructed sequence. Extensive evaluation on public data sets and comparative analysis with other occlusion handling methods verify the suitability of our approach for potential application in real-life scenarios.
%occlusion reconstruction popularly used Gait Energy Image (GEI) feature Evaluation of the algorithm The CASIA-B dataset,and TUM-KGP on which the experimental results evidently proves the effectiveness of our proposed model in handling occlusion. 
\footnotetext[1]{These authors contributed equally to this work.}
\end{abstract}


\section{Introduction}
%Gait refers to the walking pattern of an individual. 
Over the past decade, research on computer vision-based gait recognition emphasizes the fact that a gait-based biometric identification system can be potentially used to identify suspects in surveillance sites to strengthen public security. The main advantage of gait over any other biometric is that it can be captured from a distance, and the videos/images captured by surveillance camera may not be of very high resolution. Literature on gait recognition shows that, the effectiveness of most of the existing approaches \cite{han2005individual,roy2012gait,chattopadhyay2014pose} depends on the availability of %has already established the fact that accurate gait recognition can be done only if 
a complete gait cycle. % of gait is made available.
However, in most practical situations, presence of a complete cycle is not guaranteed. This is because static or dynamic occlusion can occur anytime during the video capturing phase. %Occlusion can be of two types: static and dynamic. 
In static occlusion, the occluding object (or, the occluder) remains stationary, e.g., a pillar, whereas in case of dynamic occlusion, the occluders are moving objects, e.g., walking persons. 

\begin{figure*}[!h]
\centering
     \begin{subfigure}[ ]{.95\textwidth}
      % include first image
      \centering
      \includegraphics[width=.95\linewidth]{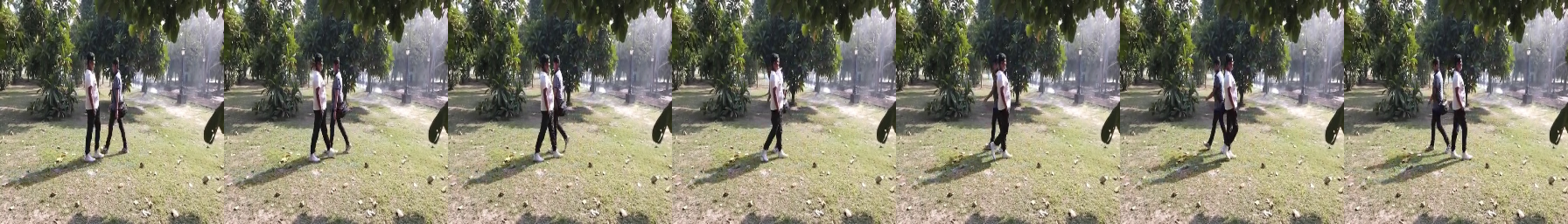}
      \caption{}
      \label{fig:res-first}
    \end{subfigure}
    % \vspace{10pt}
    % \hfill
    \begin{subfigure}[ ]{.95\textwidth}
      \centering
      % include second image
      \includegraphics[width=.95\linewidth]{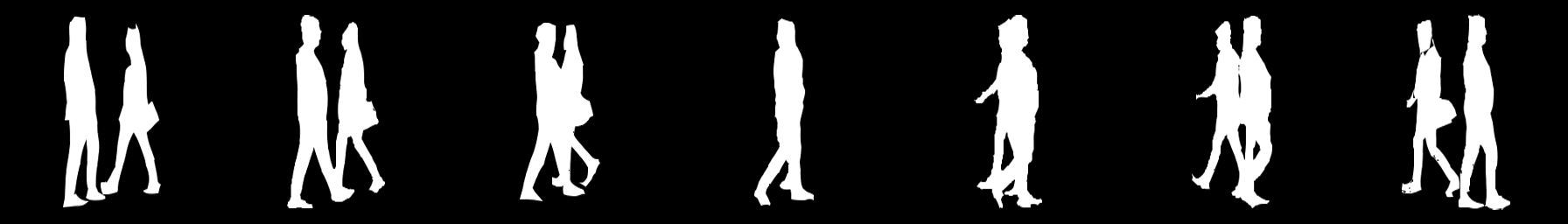}
      \caption{}
      \label{fig:res-second}
    \end{subfigure}
    % \vskip\baselineskip
    \caption{(a) An RGB sequence demonstrating dynamic occlusion and (b) Binary silhouettes extracted from the above frames}
    \label{fig:occ}
\end{figure*}

Figure \ref{fig:occ} explains a specific case of dynamic occlusion scenario in which RGB frames of a gait sequence are presented in the first row (i.e., Figure \ref{fig:occ}(a)), and the corresponding background subtracted and pre-processed frames are shown in the second row (i.e., Figure \ref{fig:occ}(b)). 
It can be clearly seen from Figure \ref{fig:occ}(b) that clean silhouette of a target subject is not continuously available after applying standard pre-processing techniques, which makes classification challenging. Previous approaches to gait recognition are not suited to handle these situations effectively. 
%in difficulty in recording a continuous gait cycle. Due to occlusion only disjoint fragments of the gait cycle of a target subject are available. 
The difficulty of the problem increases if the recorded sequence is small due to capturing only a part of a gait cycle. In that case, replacing the occluded frames of a gait cycle by matching frames from other cycle/s as described in \cite{hofmann2011identification} is also not a feasible approach. %Traditional gait recognition approaches are not suitably equipped to handle incomplete gait cycle information, and hence their effectiveness is likely to suffer in the presence of occluded sequences. Thus 
%From the above discussion, it is clear that 
Thus, for carrying out gait recognition effectively in the above-mentioned challenging situations, reconstruction of at least a complete gait cycle from the available incomplete cycle information seems to be a necessity.

Till date, only a few techniques \cite{chattopadhyay2015frontal,babaee2019person,roy2011occlusion} have been developed to perform gait recognition in the presence of occlusion. However, the results reported by these approaches are not accurate enough, and hence these are unsuitable for potential deployment in practical scenarios. We propose effective deep learning-based algorithms for detecting and reconstructing frames in a gait sequence that got corrupted due to occlusion. % in a gait cycle that are either corrupted or went missing due to the occurrence of static/dynamic occlusion. 
A VGG-16 network is initially employed to detect the occluded frames present in an input sequence. Next, reconstruction of missing silhouettes in these occluded frames is carried out by employing a recurrent deep learning model. Our approach has been seen to perform robustly against varying degrees of occlusion through extensive experimental evaluation. The only existing gait data set featuring occlusion, namely the TUM-IITKGP data set \cite{hofmann2011gait}, consists of only a small set of subjects. Hence, we consider another popular extensive gait data set of non-occluded sequences, namely the CASIA data \cite{zheng2011robust,yu2006framework}, and introduce varying degrees of synthetic occlusion in it to prepare the data for training the occlusion detection and reconstruction models. The proposed occlusion reconstruction model has been termed as \emph{Reconstruct Gait Net} (abbreviated as \emph{RGait-NET}). 

It needs to be mentioned here that the novelty of the work is developing deep learning models for occlusion detection and reconstruction. For gait feature extraction and recognition, we have used popular existing methods. Hence, discussions related to occlusion handling has been presented in a more detailed manner in the paper, and less focus have been given to discussions related to gait feature extraction and classification. The main contributions of the work may be summarized as follows:
\begin{itemize}
    
    \item To the best of our knowledge we, for the first time, are proposing neural based approaches to detect occlusion as well as predict missing frames in an occluded gait sequence consisting of a single gait cycle.
    
    \item Training a VGG-16 network to effectively distinguish between occluded and non-occluded frames, and proposing a novel LSTM-based frame reconstruction module with multi-objective loss function to train the network so as to effectively perform frame reconstruction. % frames of the gait sequences.
    
    \item Extensive experimental evaluation and comparison with state-of-the-art approaches to test the effectiveness of our approach. The trained models as well as data sets used to obtain the trained model has been made available for further comparative studies.
\end{itemize}

The rest of the paper is organized as follows. Section \ref{bg} presents the related literature focusing on occlusion handling techniques in gait recognition. Next, in Section \ref{pa}, we present the occlusion detection and reconstruction algorithms along with the network architecture details. Next, in Section \ref{ee}, the evaluation details are presented and important findings of the paper are highlighted. Section \ref{con} finally concludes the paper and points out future scopes for work.%, and the trained model can be downloaded by clicking  \href{https://drive.google.com/file/d/1jqjSb58x8yao1vOJNLJstNvxoZjjuyGg/view?ts=5dcb0388}{here}. To the best of our knowledge, ours is the first-ever work on occlusion reconstruction in gait sequences by exploiting the generalization capability of neural networks. 
%The gait recognition scenario considered in this paper can be explained with the help of Figure \ref{fig:figure1}. 

% of a person at public places seems quite illogical. So need of some non-invasion and non-cooperation based identification technique is indispensable. The manner of walking of each person is unique and can be used as a pragmatic approach for human identification. Since human movement is periodic, it is sufficient to capture one complete gait cycle to recognize the corresponding person. A gait cycle can be defined as the time period or sequence of movements during locomotion in which one foot comes in contact with the ground to when the same foot again contacts the ground. But in practical scenarios it is sometimes not possible to capture a complete or clean gait cycles, this condition is called occlusion. Occlusion can occur due to presence of multiple persons in the same image or presence of beams, pillars, and other non-living objects. Thus, person identification from occluded scenes can be performed well by the existing state-of-the-art methods on gait recognition.
%In the present paper, we focus on gait recognition from occluded gait cycles. We have proposed an end-to-end neural based approach that can perform occlusion detection and simultaneously generate the occluded frames. 

\section{Related Work}\label{bg}
Traditional approaches to gait recognition can be classified as either appearance-based \cite{han2005individual,xu2007marginal,roy2012gait, zhang2010active,collins2002silhouette, chattopadhyay2015frontal,sivapalan2011gait,chattopadhyay2014pose} or model-based \cite{Model1,Model2,Model3,3DGait,PHASEWTD}. The main difference between the two categories is that while appearance-based techniques  utilize the silhouette information from the video frames to extract features for gait recognition, the model-based methods  attempt to fit the kinematics of human motion in a pre-defined model. Computation of appearance-based gait features can be done very fast without undergoing complex model fitting operations. These have also been seen to provide significantly accurate classification results, and hence gained higher popularity compared to the model-based approaches. 
Early work on appearance-based gait recognition mostly use gait videos captured by RGB cameras and focus on the fronto-parallel, i.e., side view of gait, since binary silhouettes from the fronto-parallel view provide maximum gait information. In the later years, few cross-view gait recognition were proposed in which the training and the test sets are captured from different views. For example, the work in \cite{ben2019coupled} performs recognition by aligning the GEIs \cite{han2005individual} from different view-points using coupled bilinear discriminant projection (CBDP). On the other hand, a CNN-based architecture for cross-view gait recognition have been proposed in \cite{takemura2017input}.

With the introduction of RGB-D cameras such as Kinect, a few frontal-view gait recognition techniques \cite{sivapalan2011gait,chattopadhyay2014pose} have also been developed. Advantages of frontal view gait recognition is that it is less prone to occlusion, as a result of which there is a higher chance for capturing clean and usable gait cycle information even from a short sequence. Since, reliable gait features cannot be extracted from frontal view binary silhouette sequences, depth streams provided by depth cameras such as Kinect have been mostly utilized in research on frontal gait recognition. In \cite{battistone2019tglstm}, an LSTM-based gait recognition method is presented that uses a graph-based learning approach to capture the spatial and temporal aspects of gait in an effective manner.

The gait recognition scenarios considered by the above techniques is very simplistic in the sense that these consider only one person to be present in the field of view of a camera. However, a proper gait recognition algorithm must not be dependent on such constraints. Handling occlusion in gait recognition effectively is extremely essential since occlusion is an inevitable occurrence in any practical situation. %Occlusion while reless and until the walking sequence is recorded in suitable environment due to crowding in areas like railway station, airports, shopping malls, etc. 
%Occlusion in videos can occur in various forms,  %be of distinguished into mainly types that is e.g., self-occlusion \cite{cho2013adaptive,cho2012self} which refers to the situation when part of the body gets occluded by another part, and occlusion that are caused by other/external objects \cite{roy2011occlusion,chu2013tracking}. Occlusion can be further categorized as static occlusion, i.e., occlusion caused due to stationary objects such as pillars, %in the path of camera and the subject whose gait is being recorded and dynamic occlusion, i.e., occlusion caused due to moving objects. %The problem statement that we address includes addressing all kinds of occlusion particularly static and dynamic occlusion. It is due to occlusion in videos recorded that brings noise to the feature vector generated to be used for identification, 
Presence of occlusion makes the silhouettes in the video frames noisy, and also hinders the capturing of a complete clean gait cycle. This affects the recognition accuracy of most traditional appearance-based approaches, as discussed before. %Till date, only a few gait recognition exist in the literature that consider occlusion in gait sequences \cite{5774992,roy2011occlusion,hofmann2011gait}, but none of these approaches is accurate enough for practical deployability. 
Some recent approaches towards handling the problem of occlusion in gait recognition are discussed next.

%Initially approaches directly extracted features from Gait Energy Image and Color Histograms and used them for human identification. 
Occlusion reconstruction has been done using a Gaussian process dynamic model in %to reconstruct the occluded frames like
\cite{roy2011occlusion}. In this work, occluded frames in a gait sequence are first detected and next these occluded frames are reconstructed by assuming that the variation of gait features over a gait sequence can be modeled by fitting a multi-variate Gaussian distribution. The viability of this approach has been evaluated using the TUM-IITKGP data \cite{hofmann2011gait}. 
% In \cite{isa2010gait} an interpolation based approach using Support Vector Machine(SVM) has been adopted to reconstruct the occluded frames. For classification, dimensionality reduction based approaches like Principal Component Analysis(PCA) and Canonical Analysis(CA) have been applied on the reconstructed data. 
In \cite{isa2010gait}, the authors proposed an approach that uses Support Vector Machines-based regression to reconstruct the occluded data. %For recognition, Principal Component Analysis(PCA) and Canonical Analysis(CA) is used. 
This reconstructed data is first projected onto a PCA subspace, and next %Canonical Analysis is applied to projected data and 
classification of the projected features to the appropriate class has been carried out in this canonical subspace. Three different techniques for reconstruction of missing frames have been discussed in \cite{lee2009coping}, out of which the first approach uses interpolation of polynomials, the second one uses auto-regressive prediction, and the last one uses a method involving projection onto a convex set. In \cite{aly2014partially}, an algorithm focusing on tracking of pedestrian is presented, in which the results are evaluated on a synthetic data set containing sequences of partially occluded pedestrian. A number of approaches involving human tracking \cite{zhang2010robust,andriluka2008people} and activity recognition  \cite{de2002continuous,weinland2010making} techniques have also been developed which handle the challenging problem of occlusion detection and reconstruction. 

The work in \cite{roy2015modelling} presents modeling and characterization of occlusion in videos. The main contribution of this work is %not deriving novel gait features, but it 
providing a suitable way of evaluating the effectiveness of occlusion handling algorithms since data sets with real occlusion are scarcely available. The algorithm presented here is about introducing either static and dynamic occlusion synthetically in a non-occluded video taking into consideration a set of inputs provided by user, e.g., time of occlusion, starting pose of the target subject during occlusion, etc. Characterizing the level of occlusion in a given video is done by employing a particle swarm optimisation-based parameter estimation technique.  In \cite{babaee2018gait}, a deep neural network-based gait recognition approach has been proposed %to identify a person from an incomplete gait cycle by 
to reconstruct \emph{Gait Energy Image} from an input sequence which does not contain all the frames of a complete gait cycle. Specifically, here the authors have employed a fully convolutional Autoencoder that predicts the desired GEI feature from %takes as input the averaged silhouette corresponding to the frames of an 
an incomplete gait cycle. % and outputs the GEI \cite{han2005individual}, i.e., the averaged silhouette corresponding to a complete gait cycle. 
However, this approach has not been tested for varying levels of occlusion. Also, availability of only few frames of a gait cycle is likely to drastically affect the prediction performance of the model.
%This approach can also be used with other gait features like Gait Entropy Image, Gait Flow Image or Motion History Image, etc. 
An improvement to the above work has been proposed in \cite{babaee2019person} by the same authors in which the % of \cite{babaee2018gait} released an improved version in 
transformation from incomplete GEI obtained from the incomplete gait cycle to the complete GEI has been performed in a progressive manner. %The idea behind using progressive transformation is the unavailability of sufficient number of frames. 
%In case %there might be only a few frames of a gait cycle are available, as discussed above, %so in such cases  generating GEI corresponding to the complete cycle is not a trivial task. % large number of frame differences between the incomplete GEI and the complete GEI is not trivial. So to address 
%This problem is handled by training 
In this work, a large number of Autoencoders have been trained to predict various poses in a gait cycle, which are later combined in an end-to-end manner.

%The only data set in the public domain involving real life occlusion situations is the TUM-IITKGP data set \cite{}. 
The above methods deal with occlusion in a gait sequence in which binary silhouettes corresponding to corrupted intermediate frames of a gait cycle are reconstructed. However, there might be situations where only small fractions of gait cycle gets captured due to short length of the monitoring zone. Situations like this can be found at the security check-points in airports, railway stations and shopping malls. Gait recognition from such incomplete gait cycle information has been done in 
%Some work done in the domain of occlusion are address the problem of presence of incomplete gait sequence. In
\cite{chattopadhyay2013gait} using the depth and skeleton streams provided by Kinect. % a pose energy image based human identification approach from incomplete gait sequence of a subject has been defined. 
Here, a set of key walking poses are first constructed, and next individual frames of a gait cycle are mapped into the appropriate key poses by means of a state-transition model. The trajectory of each skeleton joint (obtained from the skeleton stream of Kinect) is used to compute features representing the dynamics of gait. Based on these features an initial level of short-listing of the gallery set is done, and next silhouette-level features extracted from the depth stream of Kinect are used for final prediction. Comparison at each level is done with respect to the available key poses present in both the training and test sequences. %The authors have taken into account the conditions in which kinect depth camera are installed on top of entry or exits. \cite{chattopadhyay2015frontal} focus on frontal gait recognition in presence of occlusion. Two kinect cameras facing each other are used to extract the front and back gait features of a person which are used to identify a test subject. Smith-Waterman algorithm has been used to to map the occluded frames in the test set to the frames in the gallery set of frames that closely resembles them. Finally the recognition is based on euclidean distance between the test subject and the subjects in the training set.
The work in \cite{chen2009frame} presents a robust gait representation technique termed as \textit{Frame Difference Energy Image} (FDEI), in which silhouette differences between successive frames are used to compute the averaged gait feature instead of the actual silhouettes. This type of feature representation technique tends to suppress the incompleteness of the binary silhouettes while simultaneously retaining the temporal and spatial information of the walking pattern. Finally, a Hidden Markov Model (HMM)-based classification approach has been employed for recognizing a test subject.

From the extensive literature survey, it is observed that most existing gait recognition approaches require a complete cycle of gait for proper functioning. Till date, no neural network based approach have been developed to detect occlusion, and predict the missing frames in an occluded sequence. The few occlusion handling strategies in gait recognition developed so far either make unrealistic assumptions such as walking pattern follows a Gaussian distribution, or are applicable only in highly constrained scenarios where multiple gait cycles must get captured.  %Also, most of these techniques  have not been tested exte their viability of their approaches on real data sets containing natural occlusions. 
In this work, we improve the state-of-the-art research on gait recognition by proposing a novel frame reconstruction algorithm. Our approach predicts the missing silhouettes in a gait sequence from only limited gait information present in a few initial non-occluded frames by employing an effective neural network-based learning algorithm. 
% Recent approaches like \cite{babaee2019person,chen2009frame}  involving deep learning addresses the problem conditioned to presence of incomplete gait cycle of any subject.
% However, these approaches do not address the problem of occlusion directly that leads to poor performance in the retrieval results.

\section{Proposed Approach}\label{pa}
%Usually if features like Gait Energy Image, Pose Energy Image, etc. are directly applied to the sequence containing occlusion the performance of the existing algorithms go down due to the redundancy in the occluded frames that create distortion in the final features leading to bad performance. Thus to this end 

\begin{figure*}[h]
     \centering
     \includegraphics[width=.9\textwidth]{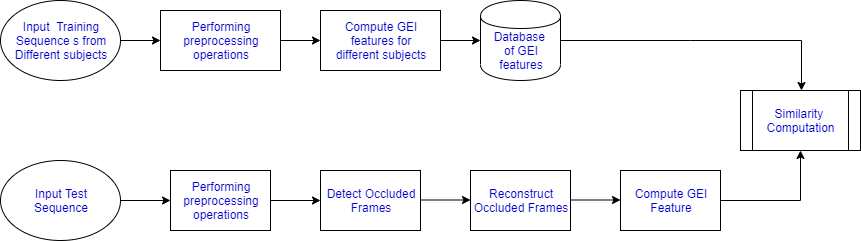}
     \caption{A block diagram explaining the pipeline of the proposed gait recognition algorithm}
     \label{fig:algofigure1}
\end{figure*}

%We propose a two stage method, mainly occlusion detection followed by occluded frame reconstruction. 
A schematic diagram explaining the steps of the proposed approach, namely the occlusion detection and occlusion reconstruction modules are shown in Figure \ref{fig:algofigure1}. 
With reference to the figure, person identification from his/her gait signature has been carried out by comparing a test subject against a gallery of a large number of subjects. At first, %the number of frames in a complete gait cycle is estimated. Next, 
each frame in a given gait sequence is classified as either occluded or non-occluded by employing a VGG-16 model, and next occluded frames in the cycle are reconstructed with the help of a generative neural network model. Finally, the Gait Energy Image (GEI) feature is computed from this reconstructed sequence, and the class of the test subject is determined by comparing the GEI feature of the test subject against a gallery of GEI features corresponding to a large number of subjects using Random Forest classifier. The steps of the proposed approach are explained in further detail next.
%to  discriminative and efficient features even from the gait sequences having occlusion. Our method is robust to any real life data-set since occlusion is detected and then we reconstruct those occluded frames to reduce redundant information in the occluded frame. Let as consider a sequence of $N$ frames $\mathcal{F}$ = \{$\mathcal{F}_1$ $\mathcal{F}_2$ ... $\mathcal{F}_N$\}. Let $G$ and $F$ be the invertible functions that predicts the probability of occlusion of $i_{th}$ frame and reconstructs the occluded frame as $\hat{\mathcal{F}_k}$ respectively.In further sections we will discuss the stages in detail. We will discuss them in the following sections.

\subsection{Occlusion Detection}\label{od}
%Initially the frames registered from a video has occluded frames which are not labelled or given to be occluded. So 
Occlusion detection is carried out in an automated manner by employing a deep VGG-16 Convolutional Neural Network. % (more specifically, a VGG-16 network). %Transfer learning which has shown commendable results with images hence we use transfer learning for this purpose . 
% \textbf{We use %a deep convolutional neural network architecture inspired from the VGG16 architecture pre-trained on image-net dataset. An artificial 
Training of the VGG-16 network is performed by constructing a data set of 1524 images consisting of 664 non-occluded silhouettes selected from both the TUM-IITKGP \cite{hofmann2011gait} and the CASIA-B \cite{zheng2011robust,yu2006framework} gait data sets, along with 860 occluded silhouettes (featuring both static and dynamic occlusion) from the TUM-IITKGP data. %The complete data set used % to train the occlusion detection model. The VGG-16 network is next trained with this data set to carry out occlusion detection in an effective manner. %This network is trained to optimize a binary cross-entropy loss function so that it can effectively distinguish between occluded and non occluded frames. 
%Let the $i^{th}$ frame of an input gait cycle be denoted by $\mathcal{F}_i$. %let $G$($\mathcal{F}_i$) represents the function learnt by the occlusion detection model. % $\mathcal{F}_i$ is occluded.
It is imperative that the final layer of this model will have two output nodes corresponding to the \emph{Occluded} and \emph{Non-occluded} classes, respectively. Let $G_1$%$\mathcal{F}_i$) 
and  $G_2$ be the functions learnt by the occlusion detection model at the two output nodes. For an input frame $\mathcal{F}_i$, let $P_1(i)$ and $P_2(i)$ denote the probability of occlusion and non-occlusion, respectively. Then,
\begin{eqnarray}\label{pcc_prob}
P_1(i)&=&\frac{G_1(\mathcal{F}_i)}{G_1(\mathcal{F}_i)+G_2(\mathcal{F}_i)},\\
P_2(i)&=&\frac{G_2(\mathcal{F}_i)}{G_1(\mathcal{F}_i)+G_2(\mathcal{F}_i)}.
\end{eqnarray}
% ($\mathcal{F}_i$) be used to denote the functions learnt at the two output nodes, respectively. Suppose for an input frame% used to denote the probability the frame $\mathcal{F}_i$ to be occluded. 
If $y_i$ denotes the ground-truth class label for $\mathcal{F}_i$ %and $G_1$($\mathcal{F}_i$) is the predicted probability by the occlusion detection model, 
then the cross-entropy loss function $L_{occ}$ with trainable parameters is computed as:
\begin{equation} \label{occlusion loss}
    %L_{occ} = -y_ilog(G(\mathcal{F}_i)) - (1-y_i)log(1-(G(\mathcal{F}_i))).
     L_{occ} = -y_ilog(P_1(i))-(1-y_i)log(P_2(i)).
\end{equation}
The VGG-16 network is trained with RMSprop optimizer for a maximum of 100 epochs or until the loss value (computed from (\ref{occlusion loss})) attains a saturation level, i.e., the loss between two successive epochs is smaller than a pre-defined threshold $\epsilon$. Usually, $\epsilon<<1$ and here we choose its value to be $2e-4$. On completion of the training phase, the model learns to accurately differentiate between occluded and non-occluded frames. %To prepare the gallery set for training the occlusion detection model, we select non-occluded frames from both the CASIA-B and the TUM-IITKGP data set, and occluded frames from the TUM-IITKGP data set has been only. 
The trained VGG-16 model %can be downloaded by clicking \href{https://drive.google.com/file/d/1U0GGlujHCCkusZkGQwuTvi5LmznhRtHI/view}{\textit{here}}, 
and the gallery set used for training the above model made available for further comparison (link given at the end of Section \ref{ee}). %\href{https://drive.google.com/drive/u/0/folders/1POkwH08NIKy8ee6VqaOqG2UtUlo7VPJj}{\textit{here}}. % as well as the trained model \footnote[3]{\url{https://drive.google.com/file/d/1U0GGlujHCCkusZkGQwuTvi5LmznhRtHI/view}} has been made available to the research community for comparative studies.

%Hence, for a test sequence we find all the occluded frames and their indices and proceed to the occlusion reconstruction part. %Thus we fine-tune the pretrained VGG-16 model in our dataset such that it can efficiently learn to classify between occluded and non-occluded frames.   %The initial 13 layers of VGG-16 network were freezed and rest of the layers were made trainable followed by a Global Average Pooling layers and 128 neurons dense layers with BatchNormalization and dropout (with dropout rate = 0.5). The last layer was single neuron with sigmoid activation function. %The whole network is combinedly trained with RMSProp as the optimizer and binary crossentropy as the loss function.

%To train the occlusion detection model such that it could accurately differentiate between occluded and non occluded frames, we created a data set on which we fine-tuned the VGG 16 model. A data set of 1524 images was created consisting  860 occluded frames out of which some were collected from the TUM- IITKGP data set and some had random noise added. Also 664 normal walking frames were collected from the TUM-IITKGP data as well as the CASIA-B Gait database to make to model's learning robust. %We trained the model for 50 epochs with RMSprop as optimizer and learning rate kept as 0.0001 until the loss saturated. The model achieved a training accuracy of $97.03\%$.

\subsection{Occlusion Reconstruction}\label{or}
Once occluded frames are detected by applying the trained VGG-16 network, %The occlusion reconstruction model % enables us to labels the occluded frames out of the frames recorded for a subject, hence we now propose 
an occlusion reconstruction model is employed to reconstruct these corrupted frames and generate clean silhouettes. Any gait sequence can be viewed as a time-series data in which the silhouette at a particular instant of time is dependent on the silhouettes preceding to it. Therefore, it appears that silhouette information embedded in the previous frames can be used to predict the subsequent missing/corrupted frames of a gait sequence. %So, the 
In this work, we propose to perform the reconstruction by means of %accumulated silhouette knowledge corresponding to the previous frames %instead of %can help in reconstructing the occluded frames with high precision instead of %using only just one previous frame as more data about the past frames which would provide the model better learning and improve the prediction capability of the is used to train using 
a deep time-series neural network, more specifically a Fully Convolutional Long-Short Term Memory (FC-LSTM) network  \cite{DBLP:journals/corr/ShiCWYWW15}. This network uses convolutional layers in place of the dense layers present in traditional LSTMs. %Although Fully-Connected LSTMs have proven powerful in handling temporal correlation, it contains too much redundancy for spatial data. Since 
In the past, FC-LSTMs have been successfully used in prediction of temporal data \cite{xingjian2015convolutional}. % and also the benefits of Convolutional Neural Network in extracting image features cannot be over-emphasized. Hence, 
Since, gait of a person also follows a specific temporal pattern, it appears that FC-LSTM can predict occluded frames effectively with the knowledge of previous un-occluded frames in a gait sequence.
%we have used the Convolutional LSTMs as proposed in \cite{} which uses Convolutional layers in place of dense layers inside the LSTM . 

%Now we know from occlusion detection that which frames of the registered gait sequence are occluded hence now we propose an occlusion reconstruction model which effectively uses the information from the previous frames which are non-occluded and predicts the occluded frame We have used Convolutional Long Short Term Memory(ConvLSTM) as the building blocks of our network architecture since LSTMs have shown good results in prediction of time series data and Convolutional Neural Networks perform well with images hence we combined both and have used ConvLSTM in our model. %For the occlusion reconstruction model 

The FC-LSTM model used in this work consists of a deep Auto-Encoder with seven convolutional layers, four time-distributed pooling layers stacked on top of each other. For regularization, we consider dropout layers with a dropout probability of 0.2, and apply batch normalization after each convolutional layer. To further improve the robustness of our model we add zero centered Gaussian noise to the input silhouette. We have also made use of skip connection in our model %that can be seen in fig\ref{fig:network}%
for better gradient flow throughout all the nodes of the graph and to minimize the effect of the vanishing gradient problem. Let the function learnt by the occlusion reconstruction model be represented by $T$ that takes as input a sequence of gait frames represented by $\mathcal{F}$ = \{$\mathcal{F}_1$ $\mathcal{F}_2$ ... $\mathcal{F}_K$\} and outputs the $\mathcal{F}_{k+1}^{th}$ frame. Mathematically, this can be represented as:
\begin{equation}
    T(\mathcal{F}_1,\mathcal{F}_2, ..., \mathcal{F}_k) = \mathcal{F}_{k+1}.
\end{equation}

%\subsection{Objective Function}
\begin{figure*}[h]
     \centering
     \includegraphics[width=\textwidth,height=7cm]{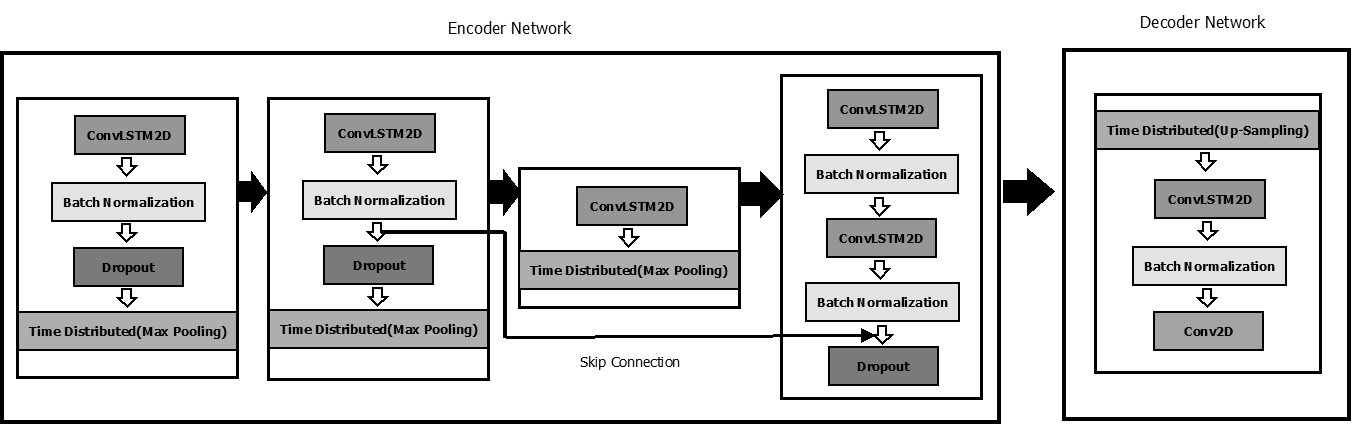}
     \caption{Architecture of the proposed occlusion reconstruction model}
     \label{fig:figure1}
\end{figure*}

We make use of a multi-objective loss function to train the FC-LSTM network %which is optimized by the 
for occlusion reconstruction. % model. Our 
%The loss function enables the network to predict proper representation of silhouette frames. The objective function 
It consists of two components, namely a binary-cross entropy loss and a dice loss that are explained next. %coefficient functions.
The silhouette images have pixel intensities of either \emph{0} or \emph{1}, %hence we incorporate the binary cross entropy loss in the form of reconstruction loss function in our model. 
and we compute the pixel-wise cross-entropy loss ($L_{rec}$) between the reconstructed image and the ground truth image as shown in (\ref{reconstuction loss}): %hence improving the reconstruction ability of the network.
\begin{align} \label{reconstuction loss}
    L_{rec} = -y_{i+1}log(T(\mathcal{F}_1,\mathcal{F}_2, ..., \mathcal{F}_i)) - \\\notag (1-y_{i+1})log(1-(T(\mathcal{F}_1,\mathcal{F}_2, ..., \mathcal{F}_i))).
\end{align}
%where the function $T$ takes initial $i$ number of frames $\mathcal{F}_1$, $\mathcal{F}_2$, ..., $\mathcal{F}_i$ and generates the $(i+1)^{th}$ frame $G(\mathcal{F}_{0-i})$ and 
In the above expression, $y_{i+1}$ denotes the ground truth frame label. %Secondly we propose to use dice loss in our loss function. It can be understood that The reconstruction ability of model can be improved if the model prediction can be improved such that silhouette that perfectly overlaps with the ground truth frame. Hence using this idea 
In addition to the binary cross-entropy loss, a dice loss has also been used to train the %to maximize the extent of overlap of object pixels between a reconstructed frame and the ground truth frame. This dice coefficient has been added as an additional loss term during training the 
occlusion reconstruction model. %to boost up the effectiveness of the reconstruction model. 
Inclusion of dice loss in the loss function helps to improve the overall reconstruction quality by maximizing %It is represented as a coefficient that gives the measure of 
the overlap between the ground truth and the reconstructed silhouettes. The dice loss provides a value % by giving values 
between \emph{0} and \emph{1}, where a value closer to 1 represents high overlap, while a value closer to \emph{0} corresponds to minimum overlap. From the above discussion, it is clear that the dice loss must be maximized for proper reconstruction. 
% Thus we aim to increase it's value in the loss function.
Hence, if $\mathcal{F}_i$ represents the actual frame and $\widehat{\mathcal{F}_i}$ denotes the reconstructed frame, then the dice loss coefficient is computed as:
\begin{equation} \label{eq3}
%\begin{split}
 L_{Dice} = \frac{2\widehat{\mathcal{F}_i}\mathcal{F}_i}{\widehat{\mathcal{F}_i}^2+\mathcal{F}_i^2}.
%\end{split}
\end{equation}
The final objective function is computed as a weighted summation of the binary cross-entropy and the dice coefficient, as shown in (\ref{eq4}). 
\begin{equation} \label{eq4}
\begin{split}
 L_{total} & = \lambda_1 L_{rec} + \lambda_2 L_{Dice}\\
\end{split}
\end{equation}
where $\lambda_1$ and $\lambda_2$ are constant  parameters. Here, the value of these parameters are set to be 1 ans -1, respectively.
%\subsection{Classification}
%A Random Forest classifier is finally employed to perform classification based on the Gait Energy Image (GEI) \cite{han2005individual} feature. For pre-processing the image size is first reduced to 128*128 and then Principal Component Analysis (PCA) is applied to the image to project it into a latent space preserving 98\% of the variance present in the training data. The random forest classifier is trained with the GEIs corresponding to the gallery subjects, and next this trained model is used to predict the class of an unknown subject.

%\subsection{Training}\subsubsection{Frame Prediction Model}
The occlusion reconstruction model is trained on the publicly available CASIA-B gait data set. This data set consists of a total of 106 subjects with six walking sequences corresponding to each person. Out of these, %in the folders namely $nm01-nm06$ out of which we use 
four walking sequences %from folders $nm01-nm04$ 
of each person have been used to train this model. % while the rest two folders are used during the testing time. 
For any sequence of the CASIA data, a set of frames from the sequence form the input to the model, and the immediate next frame in the sequence form the output. Likewise, several training sets can be formed from a single gait sequence, and these form the complete gallery data set. Training is done for a total of 62 epochs following which the training loss across epochs has been seen to attain saturation. Here, also RMSprop (Root Mean Square Propagation) has been used as the optimizer with learning rate 0.001. % and $\lambda_1$ and $\lambda_2$ as 1 and -1 respectively as mentioned in eq(\ref{eq4}). Since we need to train our model such that it can predict any pose of each subject hence during each epoch 
%the input data is formed by randomly selecting the previous number of frames to predict the next frame for each subject.

A Random Forest classifier is finally employed to perform classification based on the Gait Energy Image (GEI) \cite{han2005individual} feature. For pre-processing, the image size is first reduced to 128$\times$128 and then Principal Component Analysis (PCA) is applied to the image to project it into a latent space preserving 98\% of the variance present in the training data. The random forest classifier is trained with the GEIs corresponding to the gallery subjects, and next this trained model is used to predict the class of an unknown subject.

% \subsubsection{Occlusion detection model}\label{odm}
% To train the occlusion detection model such that it could accurately differentiate between occluded and non occluded frames, we created a data set on which we fine-tuned the VGG 16 model. A data set of 1524 images was created consisting  860 occluded frames out of which some were collected from the TUM- IITKGP data set and some had random noise added. Also 664 normal walking frames were collected from the TUM-IITKGP data as well as the CASIA-B Gait database to make to model's learning robust. We trained the model for 50 epochs with RMSprop as optimizer and learning rate kept as 0.0001 until the loss saturated. The model achieved a training accuracy of $97.03\%$.

\section{Experimental Evaluation}\label{ee}
The proposed algorithm has been implemented on a system with 96 GB RAM, one i9-18 core processor, along with three GPUs: one Titan Xp with 12 GB RAM, 12 GB frame-buffer memory and 256 MB BAR1 memory and, two GeForce GTX 1080 Ti with 11 GB RAM, 11 GB frame-buffer memory and 256 MB of BAR1 memory. 
%\subsection{Data Set Description}
%We evaluate the performance of our model on two publicly available data sets, namely the CASIA-B data \cite{} which contains occluded gait sequences only, and the TUM-IITKGP data \cite{} which contains gait videos captured under static and dynamic occlusion. %Since the proposed framework uses data from previous frames of a gait sequence to predict the next or upcoming frame of the sequence, we assume that initially few non-occluded frames are available. Since the CASIA-B Gait Data does not contain occluded sequences, 
%randomly in any frame of the sequence keeping some initial number of frames to be non-occluded. For the CASIA Gait Database, sequences in the folders $nm01,nm02,nm03,nm04$ have been used for training purpose and the sequnces in folders $nm05,nm06$ have been used for evaluation.
%During testing using the TUM-IITKGP data set, we use the data-set with occlusion. 
Evaluation of the proposed approach has been done on two public data sets, namely the CASIA-B \cite{yu2006framework} and the TUM-IITKGP \cite{hofmann2011gait} data sets. Among these, the CASIA B data consists of 124 subjects and for each of these subjects data was captured under the following conditions: (a) six sequences with normal walk (\emph{nm-01} to \emph{nm-06}), (b) two sequences with carrying bag (\emph{bg-01} and \emph{bg-02}), (c) two sequences with wearing coat (\emph{cl-01} and \emph{cl-02}). In the present set of experiments, we use the normal walking sequences (i.e., sequences \emph{nm-01} to \emph{nm-06}) captured from the fronto-parallel view. Among these, four sequences of each subject are considered for training the Random Forest with GEI features, while the remaining two are used during the testing phase, in which we introduce synthetic occlusion of varying degrees. %in t sequences to evaluate the performance . %On the other hand, 
%%%%%%%%%%%%%%%%%%%%%%%%%%%%%%%%%%%%%%%%%%%%%%%%%%%%%%%%%%%%%%%%%%%%%%%%%%%%%

The TUM-IITKGP data set, on the other hand, consists of walking videos of 35 subjects under the following conditions: (a) one video of normal walking without any occlusion, (b) one with carrying bag, (c) one with wearing gown, (d) one with static occlusion, and lastly (e) one with dynamic occlusion. %In the present set of experiments, we use the normal walking sequences as well as sequences with static occlusion. 
In the present work, we use the normal walking video to construct the gallery GEI features for the 35 subjects, whereas videos with static and dynamic occlusion are used during the testing phase. The sequences present in the TUM-IITKGP data set are sufficiently large, and we extract eight gait cycles from the normal walking video and four sequences from each of the static and dynamic occluded videos. The eight GEIs corresponding to normal walking of each of the 35 subjects results in a total of 280 GEIs, which are used for training the Random Forest classifier to recognize humans from their gait signatures. The eight occluded sequences corresponding to the videos with static and dynamic occlusion to be used during testing are labeled as \textit{Sequence 1}, \textit{Sequence 2}, ..., \textit{Sequence 8}, respectively. %and from each of these we extract 12 non-overlapping gait cycles. % into folders such that each consists of one gait cycle such that 12 gait cycles are extracted from the walking sequences per person of each type.
%The normal walking sequences are used to compute gait features and train the Random Forest classifier for gait recognition, while the occluded sequences are used during the testing phase.
%%%%%%%%%%%%%%%%%%%%%%%%%%%%%%%%%%%%%%%%%%%%%%%%%%%%%%%%%%%%%%%%%%%%%%%%%%%%%

%\subsection{Results}\label{results}
In the first experiment, we verify the effectiveness of the proposed VGG-16 occlusion detection model by evaluating its precision and recall on the training set.  %Once the indices of the occluded frames are obtained, the occlusion reconstruction model (discussed in Section \ref{or}) is employed to predict the missing frames. 
 % that has been explained in Section \ref{od}. 
To prepare the training set for occlusion detection, we randomly select % containing either static or dynamic occlusion. We observe that in total there are 
860 occluded frames (both static and dynamic) from the TUM-IITKGP data set, and 664 un-occluded frames from both the CASIA-B and the TUM-IITKGP data set. The data set for training the VGG-16 network has been made available (link given at the end of this section) for further comparison.  %\href{https://drive.google.com/drive/u/1/folders/1POkwH08NIKy8ee6VqaOqG2UtUlo7VPJj}{\textit{here}}. 
The confusion matrix obtained after completion of training the model is shown in Table \ref{tab:em1}.
% A confusion matrix computed from the results obtained by classifying each frame of the TUM-IITKGP data set as either occluded or non-occluded using the proposed occlusion detection model is shown next. 
\begin{table}[h]
    \centering
    \caption{Confusion matrix computed from the occlusion detection results}
    \begin{tabular}{|c|c|c|} 
        \hline
        \diagbox{Predicted}{Actual}&Occluded&Un-occluded\\
        \hline
        Occluded&849&6\\
        \hline
        Un-occluded&11&658\\
        \hline
    \end{tabular}
    \label{tab:em1}
\end{table}
With reference to the confusion matrix, considering the \emph{Occluded} class as the \emph{Positive} class, and the \emph{Un-occluded} class as the negative class, it can be seen that the number of true positives, false positives, true negatives and false negatives are respectively 849, 6, 658, and 11. Thus, the precision and recall for occlusion detection are 99.53\% and 98.72\%, respectively and the overall accuracy obtained after occlusion detection is 98.89\%. %From the above results, the occlusion detection model can be said to perform quite accurately. %Since, presence of occlusion distorts the silhouette shapes to a certain extent, which also drastically affects the gait recognition accuracy, a small false negative error rate is generally more desirable than a small false positive error rate. 
It is seen that out of the 860 total number of occluded frames present in the gallery set for occlusion detection, only a small fraction (i.e., 1.28\%) has incorrectly been classified as non-occluded. This low false-negative rate has resulted in a high recall of 98.72\%, and this trained model is used for detecting occluded frames during the testing phase. %Since, synthetic occlusion has been introduced in the CASIA-B data by blackening the frames, we have not repeated the above experiment for this data set. Also, since the TUM-IITKGP data set does not provide ground-truth regarding whether a frame is occluded or not, we could not compute the above metrics for the test cases.

%Now, given an input test sequence, each frame present in the sequence is classified as either occluded or non-occluded by making use of the occlusion frame detection VGG-16 model (discussed in Section \ref{od}). 

\begin{figure}[h]
\centering
     \begin{subfigure}[ ]{.48\textwidth}
      % include first image
      \centering
      \includegraphics[width=.95\linewidth]{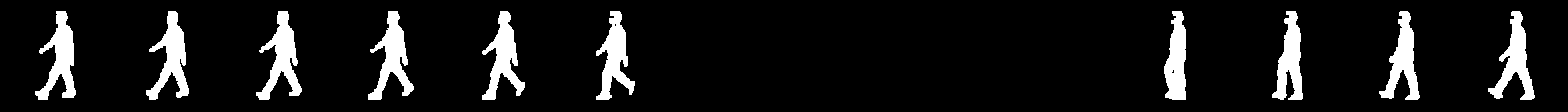}
      \caption{}
      \label{fig:res-first}
    \end{subfigure}
    % \vspace{10pt}
    % \hfill
    \begin{subfigure}[ ]{.48\textwidth}
      \centering
      % include second image
      \includegraphics[width=.95\linewidth]{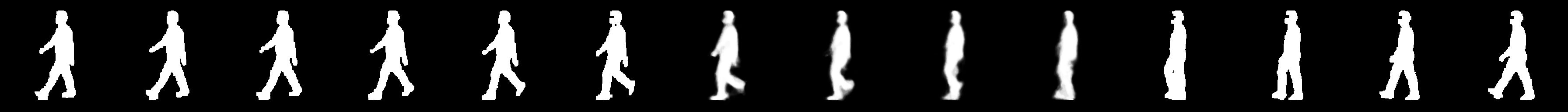}
      \caption{}
      \label{fig:res-second}
    \end{subfigure}
    % \vskip\baselineskip
    \begin{subfigure}[ ]{.48\textwidth}
      \centering
      % include second image
      \includegraphics[width=.95\linewidth]{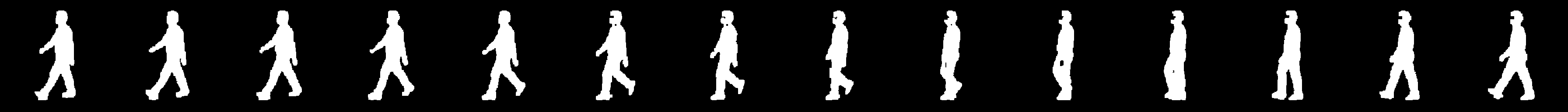}  
      \caption{}
      \label{fig:res-third}
    \end{subfigure}
    % \vskip\baselineskip
    % \FloatBarrier
    \caption{(a) Sequence with occluded frames, (b) reconstructed frames, and (c) ground truth frames}
    \label{fig:fig_res}
\end{figure}

The RGait-NET model is trained on %, as test set we first use the %original TUM-IITKGP data \cite{hofmann2011gait} which consists of sequences both with and without occlusion and is available in the public domain, and the popularly used 
the CASIA-B data \cite{yu2006framework} with synthetically introduced occlusion in its gait sequences. %Since, the original version of the CASIA-B data consists of un-occluded sequences only, we introduce different degrees of 
Synthetic occlusion is introduced in the CASIA-B data 
%The TUM-IITKGP data is available publicly, and the data set constructed by synthetically incorporating occlusion in the CASIA-B data is made available for comparative studies. 
%formed by introducing synthetic occlusion in the un-occluded sequences.
%Synthetic occlusion is imposed on the CASIA-B data 
by selecting frames at random from a binary silhouette sequence and blackening these frames (i.e., by removing the silhouette information completely). Appropriate loss functions are employed to train the network based on the differences between the  binary silhouettes produced by the LSTM-based generator and the original ground truth silhouettes (as already discussed in Section \ref{or}). One sample synthetically occluded sequence created from Figure \ref{fig:fig_res}(c) is shown in Figure \ref{fig:fig_res}(a). Each of the following experiments deal with performance evaluation of the proposed occlusion reconstruction model (RGait-NET). %The first of these is a visual inspection of the performance obtained from the proposed RGait-NET model. 
Figure \ref{fig:fig_res}(b) shows a sample output from this model on providing the binary silhouette sequence of Figure \ref{fig:fig_res}(a) as input to the network along with the frame labels (i.e., occluded or un-occluded). %The CASIA-B data has been used to train the RGait-Net model. The data set for training is constructed by selecting a large number of un-occluded sequences from the CASIA-B data, and introducing varying degrees of occlusion synthetically on these sequences. 
%Figures \ref{fig:fig_res}(a), (b), and (c) respectively show a synthetically occluded sequence from the CASIA-B data, the reconstructed sequence predicted by the proposed model, and the corresponding ground truth sequence without occlusion. 
%The synthetically occluded sequence (as shown in Figure \ref{fig:fig_res}(a)) is obtained by blackening four frames of the original binary sequence shown in Figure \ref{fig:fig_res}(c). 
Visual inspection of the images shown in Figure \ref{fig:fig_res}(b) and the ground-truth in Figure \ref{fig:fig_res}(c) verifies the fact that the reconstruction quality of the RGait-NET model (refer to Figure \ref{fig:fig_res}(b)) is indeed of high quality. %The code to generate synthetic occlusion in the CASIA-B data as well as the trained model has been made available for further comparison. 
Good quality of frame reconstruction has been obtained for the TUM-IITKGP data as well, but since this data set features real occlusion only, visual comparison with ground-truth could not be made like CASIA-B data. 
As discussed in Section \ref{or}, %the occlusion reconstruction model needs a few good quality frames at the start of the sequence to predict the reconstructed frames corresponding to the occluded frames in the later part of the sequence. 
the occlusion reconstruction model takes as input a few initial un-occluded frames to learn the walking pattern of a person using which it predicts the probable silhouette shapes in the occluded frames of the same sequence. The next experiment deals with a study of to what extent the performance of RGait-NET is dependent on the number of initial clean/un-occluded frames.  %observe the effectiveness of our approach by varying the number of initial non-occluded frames %, and also by varying the degree of occlusion in a gait sequence 
%of the sequences present in the CASIA-B data.
Results are shown in Tables \ref{tab:em1} and \ref{tab:em2} by means of Rank 1 accuracy for varying degrees of synthetic occlusion in the sequences \textit{nm-05} and \textit{nm-06} of the CASIA-B data. The initial number of un-occluded frames for obtaining the results of the two tables are considered as ten and five, respectively. %Degree of occlusion is defined as the percentage of occluded frames present in a gait cycle. 
In each of the tables, Rank 1 accuracy values are presented by grouping the percentage of occluded frames into the following five categories: 0-10\%, 10-20\%, 20-30\%, 30-40\%, and 40-50\%. The same trained model, as in the previous experiment, has also been used to obtain the reconstructed frames in this experiment. Next, GEI feature \cite{han2005individual} based human classification is performed from this reconstructed gait sequence. % obtained after passing the synthetically occluded sequences of the CASIA-B data through the RGait-NET. 
The GEI features corresponding to the walking sequences \textit{nm-01} to \textit{nm-04} present in the CASIA-B data form the gallery feature set for gait recognition for this experiment. %Testing is done by introducing synthetic occlusion randomly on the .
\begin{table}[h]
    \centering
    \caption{Rank 1 accuracy with different levels of synthetic occlusion in CASIA-B data set with ten initial non occluded frames}
    \begin{tabular}{|c|c|c|} 
        \hline
        \multirow{1}{*}{\textbf{Occlusion Degree (\%)}}  &  \multirow{1}{*}{\textbf{Rank 1 Accuracy (\%)}}\\ 
        \hline
        {$\leq$10\%}&93.39\\
        \hline
        {10\%-20\%}&86.66\\
        \hline
        {20\%-30\%}&76.66\\
        \hline
        {30\%-40\%}&63.80\\
        \hline
        {40\%-50\%}&47.37\\
        \hline
    \end{tabular}
    \label{tab:em1}
\end{table}

\begin{table}[h]
    \centering
    \caption{Rank 1 accuracy with different levels of synthetic occlusion in CASIA-B test data set with five initial non occluded frames}
    \begin{tabular}{|c|c|c|} 
        \hline
        \multirow{1}{*}{\textbf{Occlusion Degree (\%)}}  &  \multirow{1}{*}{\textbf{Rank-1 Accuracy (\%)}}\\ 
        \hline
        {$<$10\%}&92.38\\%75\\
        \hline
        {10\%-20\%}&85.71\\
        \hline
        {20\%-30\%}&73.33\\
        \hline
        {30\%-40\%}&62.85\\
        \hline
        {40\%-50\%}&40.75\\
        \hline
    \end{tabular}
    \label{tab:em2}
\end{table}
The Rank 1 accuracy reported in Tables \ref{tab:em1} and \ref{tab:em2} are the classification results obtained from the Random Forest classifier. Before using this classifier for testing, it is first tuned using the TUM-IITKGP data through cross-validation to decide upon the optimal number of decision trees in the ensemble. 
%TTUM-IITITKGP KGP data consists of un-occluded sequences as well sequences with static and dynamic occlusion for a total of 35 different subjects. %The training and test data sets for the TUM-IITKGP data we extract all frames from all the 35 videos and folders per person are created in which all the frames are stored per subject. 
From the un-occluded normal walking sequences present in the TUM-IITKGP data, we extract eight gait cycles corresponding to each subject and compute their GEI \cite{han2005individual}. Thus, a total of 280 GEIs are obtained from the gait cycles of the 35 subjects which are next used for deciding the number of decision tree predictors in the Random Forest. %, and these are used to form the training features for gait recognition. 
%Further we divide the sequence of each subject into 8 parts each part containing one gait cycle of the subject. So for the TUM-IITKGP data set we have final data-set with 35 folders corresponding to each subject with 8 sub-folders for each person with one gait cycle in each of these sub-folders. For training purpose we manually remove all the occluded frames from sequences of each subject and create GEIs for each subject. Finally for training we have 280 GEIs that is 8 GEIs per subject which are used for training the Random forest model. We project 
Before training, first, the GEIs are projected into a latent space by applying Principal Component Analysis (PCA) preserving at least 98\% variation present in the data. Random Forest with bagging algorithm is then trained on the PCA-reduced features by splitting the data into two parts: 75\% for training, and 25\% for validation. We perform this splitting ten different times randomly, and compute the cross-validation accuracy from these ten runs. The same experiment is repeated with different number of decision tree estimators: 50, 100, 150 and 200. Each of the above configurations yields a cross-validation accuracy greater than 95\%, but for 100 number of estimators a cross-validation accuracy of %100\% on training data and 
97.14\% accuracy has been obtained, which is the best among the other configurations used in the study. It may be noted that the Random Forest has been tuned using the TUM-IITKGP data set only. Although the optimal number of decision trees is expected to vary for a different data set, in this paper we use 100 trees to present the results for all future experiments on both the CASIA-B and the TUM-IITKGP data sets. This is because the main theme of the paper is to introduce deep learning-based models for occlusion detection and reconstruction. A better gait recognition accuracy may be obtained by tuning the Random Forest classifier separately for each data set, and also by using a more improved gait feature than GEI. These may be considered as future scopes of work related to occlusion handling in gait recognition.

%we primarily follow two procedure for testing purpose. Firstly we vary the amount of occlusion in the sequences to see the performance of our model and secondly we also vary the number of initial frames that are to be fed to the model. For the former setting we vary occlusion amounts from $0\%$ to $50\%$ in intervals of 10\%. %in the CASIA data-set since it has no occlusion. The results between the intervals as in table \ref{tab:em1} and table \ref{tab:em2} are averaged and written. %The first experiment was done on $nm05$ and $nm06$ in which $\%$ of occlusion is varied from $5\%$ to $50\%$ with initial non occluded frames as 5 and 10 respectively. As evident from table\ref{tab:em2} and table\ref{tab:em3}, increasing the number of initial non-occluded frames increases the rank 1 accuracy which can be justified by the fact that more number of initial non-occluded frames gives extra amount of information to frame prediction model.

%To occlude frames of the sequences in CASIA data-set we randomly select frames keeping in mind the percentage of occluded frames and add random noise to these randomly selected frames. 
%Results in Table \ref{tab:em1} represent the Rank-1 accuracy of retrieval for varying degrees of occlusion when the first 10 frames of the sequence are kept to be non-occluded. 
It is observed from Table \ref{tab:em1} that the gait recognition accuracy is quite high ($>$85\%) when the percentage of occlusion is less than 20\%. However, as we keep on increasing the degree of occlusion, the Rank 1 accuracy decreases gradually, as expected. A similar observation also follows from Table \ref{tab:em2} which presents the Rank-1 accuracy for varying degrees of occlusion if clean silhouettes corresponding to only the first five frames of the sequence are available. %Similarly in this case also rank-1 accuracy should deceases with as we increase the amount of occlusion. 
An important observation from Tables \ref{tab:em1} and \ref{tab:em2} is that the Rank 1 accuracy corresponding to the different degrees of occlusion in the two tables are closely similar to each other, which proves that the proposed RGait-NET model works satisfactorily even if a small number of initial un-occluded frames are present in an unseen gait sequence. Also its performance %with  reasonably high accuracy even if only a few clean initial frames of a sequence are available and the 
is quite reliable if the degree of occlusion is between 0-20\%.

%that on decreasing the initial non-occluded frames from 10 to 5 the accuracy of the model decreases very slightly which shows the effectiveness of our model in case when very less  number of initial frames are present. The results of Table \ref{tab:em1} and table \ref{tab:em1} are obtained using the CASIA-B Gait data-set
\begin{figure*}[h]
    \centering
    \begin{subfigure}[ ]{0.46\textwidth}
        \centering
        \includegraphics[width=\textwidth]{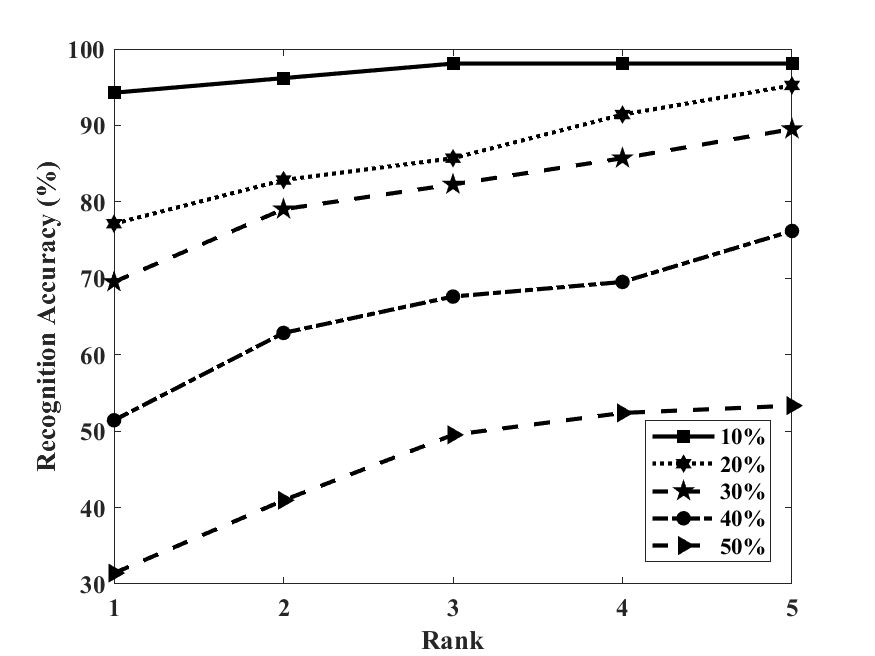}
        \caption{Sequence in folder nm-05 with 5 initial non-occluded frames}
        \label{fig:sub-first}
    \end{subfigure}
    %\hfill
    \begin{subfigure}[ ]{0.46\textwidth}  
        \centering 
        \includegraphics[width=\textwidth]{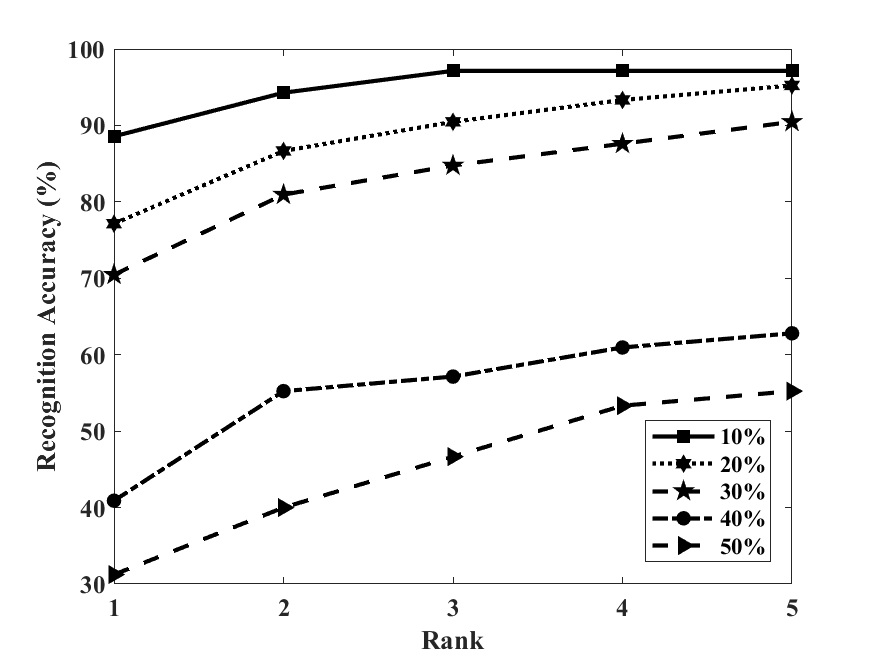}
        \caption{Sequence in folder nm-05 with 10 initial non-occluded frames}
        \label{fig:sub-second}
    \end{subfigure}\\
    %\vskip\baselineskip
    \begin{subfigure}[ ]{0.46\textwidth}   
        \centering 
        \includegraphics[width=\textwidth]{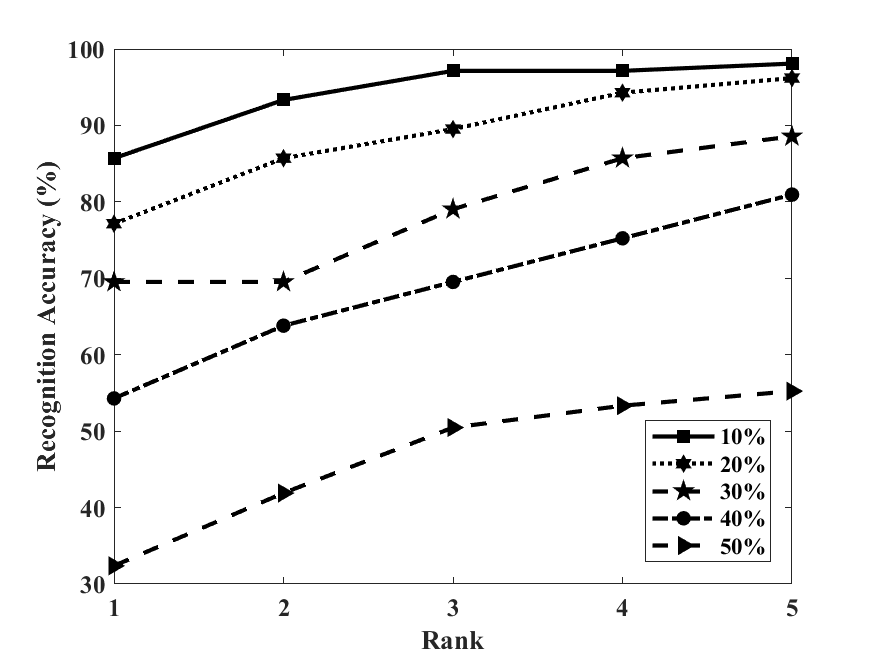}
        \caption{Sequence in folder nm-06 with 5 initial non-occluded frames}
        \label{fig:sub-second__}
    \end{subfigure}
    %\quad
    \begin{subfigure}[ ]{0.46\textwidth}   
        \centering 
        \includegraphics[width=\textwidth]{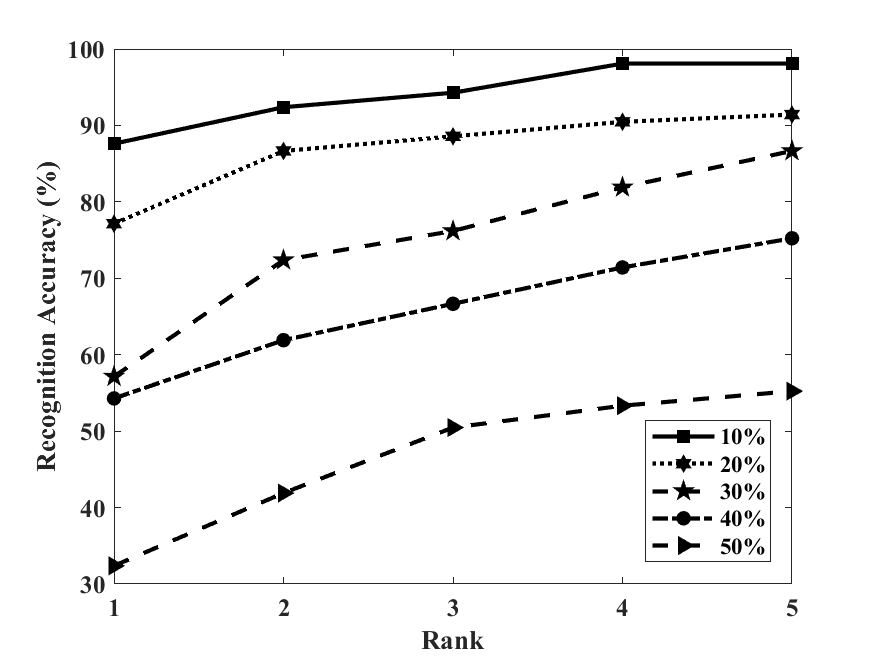}
        \caption{Sequence in folder nm-06 with 10 initial non-occluded frames}
        \label{fig:sub-second___}
    \end{subfigure}
    \caption{Cumulative match characteristic curves showing improvement in recognition accuracy with rank for the synthetically occluded CASIA-B data set}
    \label{fig:fig_}
\end{figure*}
The Rank 1 accuracy (as computed in the previous experiment) is not always a reliable metric to evaluate a classifier's performance. Rather, the improvement in recognition accuracy with increment in rank by means of Cumulative Match Characteristic (CMC) curves provides important information regarding the effectiveness of the extracted features as well as the classification algorithm. In Figures \ref{fig:fig_}(a)-(d), we plot the rank-wise improvement in the accuracy by varying the amount of synthetic occlusion in the sequences \textit{nm-05} and \textit{nm-06} of the CASIA data set. Specifically, Figures \ref{fig:fig_}(a) and (b) show the CMC curves corresponding to the synthetically occluded \textit{nm-05} sequence by setting the number of initial non-occluded frames to 5 and 10, respectively. Figures \ref{fig:fig_}(c) and (d) show the corresponding results for the synthetically occluded \emph{nm-06} sequence. % are represented through the Cumulative Match Characteristic (CMC) Curve that can be viewed in Figure \ref{fig:fig_}. 
It can be observed from the figures that our method performs
significantly accurately (greater than 85\%) if the percentage of occluded frames in a gait cycle is above 20, and it performs decently (with accuracy of more than 60\% in all cases) if the percentage is less than 30. However, as the degree of occlusion exceeds 40\%, performance of our model degrades (as expected) since in such cases more than half of a gait cycle needs to be reconstructed artificially, which also affects the gait recognition accuracy drastically. %half of the gait cycle is occluded and our model face difficulty in reconstructing frames with only half gait cycle remaining so the performance drops down.

\begin{figure}[h]
    \centering
    \includegraphics[width = 0.95\linewidth]{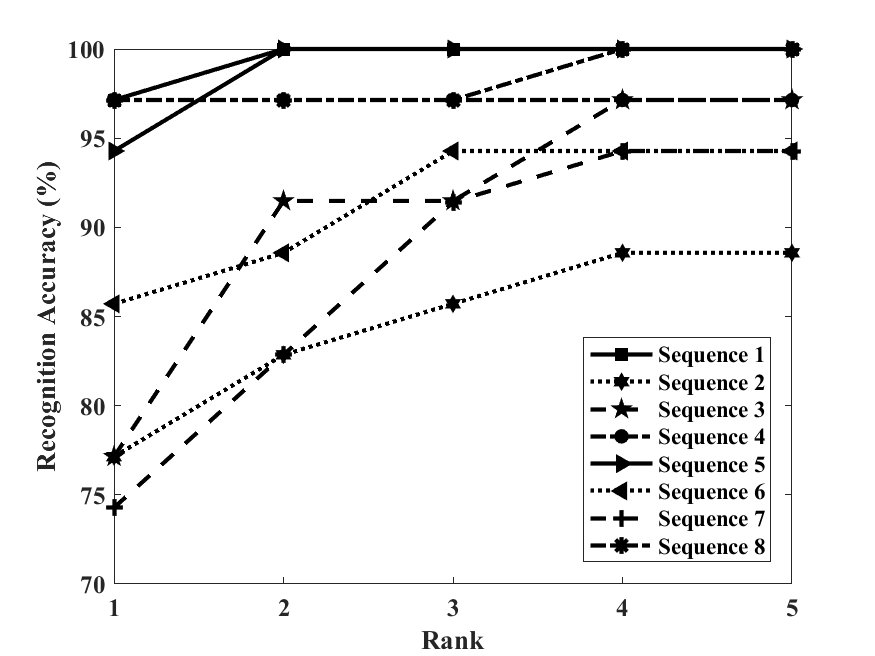}
    \caption{Rank-wise improvement in accuracy of our algorithm on the eight occluded sequences present in the TUM-IITKGP data set considering eight normal walking sequences for training the Random Forest}
    \label{tab:em3}
\end{figure}

We also perform a similar experiment using the TUM-IITKGP data. The eight GEIs computed from the un-occluded normal walking sequences are used to train the Random Forest classifier with 100 decision trees. Each of the occluded sequences extracted from the data set, namely \textit{Sequence 1}, \textit{Sequence 2}, ..., \textit{Sequence 8} are used during testing. First occlusion detection and reconstruction are performed based on the proposed deep learning models, and GEIs computed from these reconstructed sequences are input to the trained Random Forest model to determine their appropriate class.  %test sequence is next classified into the appropriate class using the Random Forest classifier based on the reconstructed GEI.
Figure \ref{tab:em3} presents a rank-wise improvement in the classification accuracy of the eight test occluded sequences by means of CMC curves as the value of the rank is increased from \textit{1} to \textit{5}. 
%For the TUM-IITKGP data-set we evaluate our model's performance on all the folders after reconstruction of the occluded frames which can be seen from table \ref{tab:em1}. 
It is seen from the figure that %the rank-wise accuracy is seen to increase for all cases. Also since the TUM-IITKGP data-set addresses real life occlusion scenarios it can be seen from table  that 
the Rank 1 accuracy of recognition is greater than or equal to 75\% for each of the sequences and above Rank 2, the accuracy is more than 80\% for all sequences.  %accuracy in all cases for rank 2 and above. %Although the TUM-IITKGP data-set contains occlusion from real life scenarios, our model still performs  well giving an
Also, the average Rank 1 accuracy over the eight sequences is 87.50\%, while the average Rank 5 accuracy over the same sequences is 96.43\%, which is significant.  %91.42\%, which can be said to be significantly accurate.

\begin{figure}[h]
    \centering
    \includegraphics[width = 0.95\linewidth]{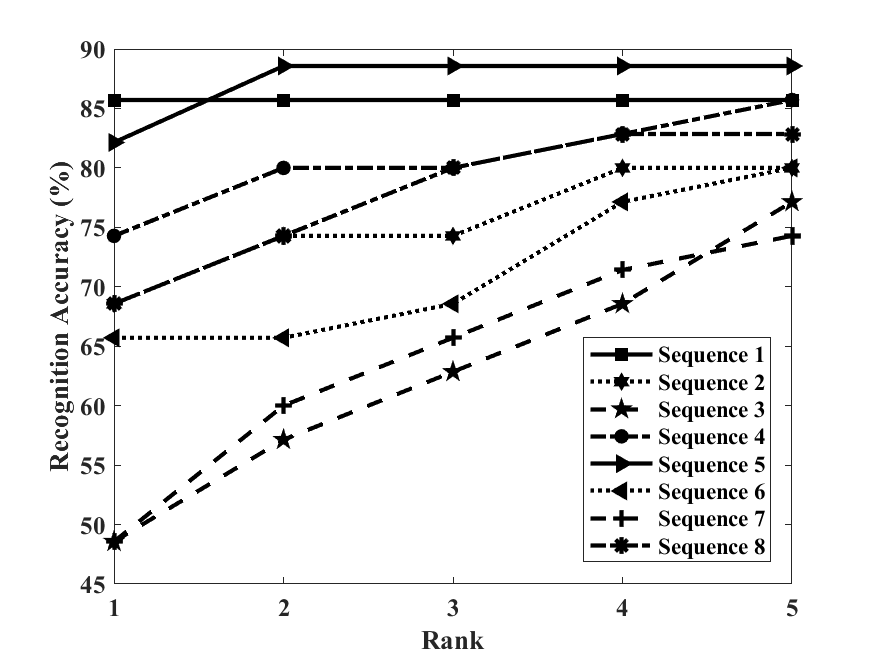}
    \caption{Rank-wise improvement in accuracy of our algorithm on the eight occluded sequences present in the TUM-IITKGP data set considering four normal walking sequences for training the Random Forest}
    \label{fig:em5}
\end{figure}

To evaluate the impact of the volume of training data on the gait recognition accuracy, we repeat the above experiment by training the same Random Forest classifier with four GEIs corresponding to each subject instead of eight (as in the previous experiment). Results are shown in Figure \ref{fig:em5} by means of CMC curves. As expected, on comparing the CMC curves of Figures \ref{tab:em3} and \ref{fig:em5}, we observe that the recognition accuracy for the different rank values reduce if the volume of the training data is decreased. This happens since presence of less training data overfits the Random Forest classifier. Still out of 35 subjects, we obtain a Rank 5 accuracy greater than 70\% for all the eight test sequences. Also, the average accuracy at Rank 1 for this particular case is 67.77\% while the average accuracy at Rank 5 is 81.78\%. 
%considering the rank 1 accuracy of all the folders. Also in the next section we compare the results of our model to other models.
%%%%%%%%%%%%%%%%%%%%%%%%%%%%%%%%%%%%%%%%%%%%%%%%%%%%%%%%%%%%%%%%%%%%%%%%%%%%
%The testing data in this case is the TUM-IITKGP dataset which were divided into 12 walking sequences out of which the occluded frames were detected using the VGG model and then were reconstructed. After reconstructing the 280 GEIs created were compiled to form the test set and were compared to the 280 GEIs obtained from the  sequences.video 1tai -> 12 normal sequence 
%%%%%%%%%%%%%%%%%%%%%%%%%%%%%%%%%%%%%%%%%%%%%%%%%%%%%%%%%%%%%%%%%%%%%%%%%%%%%
% \begin{table}[H]
%     \centering
%     \caption{Rank Wise accuracy of our algorithms on the different sequences present in the TUM-KGP data set}
%     \begin{tabular}{|c|c|c|c|c|c|} 
%         \hline
%         \multirow{1}{*}{\textbf{Folder}}  &  \multirow{1}{*}{\textbf{Rank 1}}   & \multirow{1}{*}{\textbf{Rank 2}} & \multirow{1}{*}{\textbf{Rank 3}} & \multirow{1}{*}{\textbf{Rank 4}} & \multirow{1}{*}{\textbf{Rank 5}}\\ 
%         \hline
%         1&97.14&100&100&100&100\\
%         \hline
%         2&77.14&82.85&85.71&88.57&88.57\\
%         \hline
%         3&77.14&91.48&91.48&97.14&97.14\\
%         \hline
%         4&97.14&97.14&97.14&97.14&97.14\\
%         \hline
%         5&94.28&100&100&100&100\\
%         \hline
%         6&85.71&88.57&94.28&94.28&94.28\\
%         \hline
%         7&74.28&82.85&91.42&94.28&94.28\\
%         \hline
%         8&97.14&97.14&97.14&100&100\\
%         \hline
%     \end{tabular}
%     \label{tab:em3}
% \end{table}

% \begin{figure}[H]
%      \begin{subfigure}[b]{.5\textwidth}
%       % include first image
%       \centering
%       \includegraphics[width=.8\linewidth]{nm05_5.png}  
%       \caption{NM05 with 5 initial non-occluded frames}
%       \label{fig:sub-first}
%     \end{subfigure}%
%     % \vspace{10pt}
%     \hfill
%     \begin{subfigure}[b]{.5\textwidth}
%       \centering
%       % include second image
%       \includegraphics[width=.8\linewidth]{nm05_10.png}  
%       \caption{NM05 with 10 initial non-occluded frames}
%       \label{fig:sub-second}
%     \end{subfigure}
%     \vskip\baselineskip
%     % \FloatBarrier

%     \begin{subfigure}[b]{.5\textwidth}
%       \centering
%       % include second image
%       \includegraphics[width=.8\linewidth]{nm06_5.png}  
%       \caption{NM06 with 5 initial non-occluded frames}
%       \label{fig:sub-second___}
%     \end{subfigure}
%     \quad
%     % \vspace{2pt}
    
%     \begin{subfigure}[b]{.5\textwidth}
%       \centering
%       % include second image
%       \includegraphics[width=.8\columnwidth]{nm06_10.png}  
%       \caption{NM06 with 10 initial non-occluded frames}
%       \label{fig:sub-second__}
%     \end{subfigure}
%     \caption{Rank wise CMC curve}
%     \label{fig:fig_}
% \end{figure}

%\subsection{Comparison}
%As already mentioned in Section \ref{bg}, there exist only a few work on occlusion handling in gait recognition. 
The next experiment deals with a comparative analysis of our work with that of the existing approaches on gait recognition in the presence of occlusion. We compare our work with that of \cite{roy2011occlusion,babaee2018gait}. The work in \cite{chattopadhyay2015frontal}, which describes an effective technique for occlusion handling in frontal gait recognition,  has not been used in this study since it requires the availability of the depth and skeleton streams provided by Kinect. %These methods either handle the problem occlusion (i.e., missing frames in a gait cycle) or which address the problem of unavailability of full gait cycle. 
However, to study if the proposed RGait-NET based frame reconstruction helps in generating more effective GEI features compared to the un-reconstructed sequence, we have used the GEI feature in this comparative study as well. Results are reported in Table \ref{tab:em5} both in terms of recognition accuracy and response time (i.e., the average time required to compare a pair of training and test subjects including the time for occlusion detection and reconstruction).  %baseline as we primarily use GEI as predominant attribute for feature extraction and classification. The performance of the models with our proposed model can be compared as from the 
The first column of the table corresponds to the citation of the gait recognition approach, while the second and third columns present the recognition accuracy and the response time, respectively for each approach. The Random Forest classifier trained on the eight GEIs and the same test set as in the previous experiment have been used to obtain the results for this table. 
\begin{table}[h]
    \centering
    \caption{Comparative analysis of the proposed work with existing approaches on the TUM-IITKGP data set}
    \begin{tabular}{|c|c|c|} 
        \hline
        \multirow{1}{*}{\textbf{Method}}  &  \multirow{1}{*}{\textbf{Accuracy (\%)}}   & \multirow{1}{*}{\textbf{Time (secs)}} \\ 
        \hline
        {\cite{han2005individual}}&65.86&0.09\\
        \hline
        {\cite{roy2011occlusion}}&10.25&3.65\\
        \hline
       % {\cite{chattopadhyay2015frontal}}&7.00&0.61\\
    %    \hline
        {\cite{babaee2018gait}}&34.28&0.02\\%3\\
        \hline
        {Proposed RGait-NET}&\textbf{91.42}&0.44\\
        \hline
    \end{tabular}
    \label{tab:em5}
\end{table}
It is seen from the table that the proposed approach outperforms each of the other state-of-the-art gait recognition approaches in terms of recognition accuracy by a significantly high margin. The work that is closest to our approach is the one in \cite{han2005individual} which was originally developed for un-occluded gait sequences. The approach in \cite{roy2011occlusion} fails to perform well since it works with an inherent assumption that the variation of silhouette features over a gait cycle follows Gaussian distribution, which is not necessarily true always.  Specifically, if the degree of occlusion is very high, this model is unable to estimate a proper Gaussian fit to the above feature space due to which its prediction capability suffers. The work in \cite{babaee2018gait} attempts to predict the complete GEI of a subject from the incomplete GEIs obtained from the available frames in an occluded sequence by employing a deep learning-based approach. Such an approach fails to perform well if some of the available unoccluded frames are also noisy. This is due to the fact that presence of even a few noisy frames makes the GEI from the incomplete sequence noisy due to which the quality of the reconstructed GEIs also suffers.  %assumes the availability of multiple occluded gait cycles from which it selects appropriate frames to reconstruct a complete gait cycle. The approach fails to perform well in cases where only a single occluded sequence is available, as considered in the present work. 
In contrast, reconstruction of missing frames using an LSTM-based model, as done in the present work, eliminates the above drawback to a certain extent since it takes as input individual unoccluded frames and not a single averaged frame. Also, our approach does not assume any specific distribution of human walking pattern. Using the training set for occlusion reconstruction, our approach employs a deep time-series LSTM network to model human walking as a time-series data, and automatically predict subsequent frames given a set of initial unoccluded frames.  %the silhouettes in the missing/occluded frames of a given sequence. 
The high accuracy obtained on the challenging TUM-IITKGP data set, as seen in Table \ref{tab:em5}, is due to the strong generalization capability of generative neural networks. With reference to the table, although the response time of our approach is slightly higher than that of \cite{han2005individual} and \cite{babaee2018gait}, in terms of Rank-1 recognition accuracy, our approach outperforms \cite{han2005individual} by about 25\% and \cite{babaee2018gait} by about 57\%. 

%\subsection{Model Robustness}
Finally, we perform an experiment for evaluating the robustness of the proposed RGait-NET model against different initialization of the input free network parameters. For this, we train four separate RGait-NET models from scratch using the same training set as discussed before, each with a random initialization of parameters. Next, we compute the mean and standard deviation of the recognition accuracy (for Ranks 1 to 5) for each model using the eight test sequences of the TUM-IITKGP data. Results are shown in Table \ref{tab:em4}. 
\begin{table}[h]
    \centering
    \caption{Rank-wise mean and standard deviation of the recognition accuracy obtained from the four RGait-NET models trained with different parameter initialization}
    \begin{tabular}{|c|c|c|} 
        \hline
        \multirow{1}{*}{\textbf{Rank}}  &  \multirow{1}{*}{\textbf{Mean Accuracy (\%)}}   & \multirow{1}{*}{\textbf{StdDev}} \\ 
        \hline
        % Rank 1&94.28&3.5027\\
        Rank 1&94.28&3.50\\
        \hline
        % Rank 2&95.98&2.0091\\
        Rank 2&95.98&2.01\\
        \hline
        %Rank 3&96.515&1.0825\\
        Rank 3&96.52&1.08\\
        \hline
%        Rank 4&96.96&0.3117\\
        Rank 4&96.96&0.31\\
        \hline
        % Rank 5&97.675&1.3741\\
        Rank 5&97.68&1.37\\
        \hline
    \end{tabular}
    \label{tab:em4}
\end{table}
We observe that the average accuracy of the proposed RGait-NET model is quite satisfactory for Rank 1 (i.e., 94.28\%), and for Rank 2 it achieves an almost 96\% accuracy mark, which is significantly high. Also the low standard deviation for each of the different rank values, shown in the third column of the table, indicates that the model is robust against different initial setting of the network parameters.
The data set for training the occlusion detection model, the python code to incorporate synthetic occlusion in the sequences of the CASIA-B data, as well as the trained models for occlusion detection and reconstruction have been made available in the link given in the footnote\footnote{https://sites.google.com/site/gaitrecognitioninocclusion/home}.

\section{Conclusions and Future Work}\label{con}
In this work, we have proposed a network (namely, the RGait-NET) to effectively reconstruct missing frames in an occluded gait sequence. The proposed model can be conveniently integrated with any gait recognition model to improve its accuracy on occluded sequences. We made use of two Artificial Neural Network models, one for occlusion detection, and the other for missing frame prediction, both of which take advantage of the high prediction and generalization capability of deep neural networks. The occlusion detection module is based on the popular VGG-16 architecture, and it has been trained to detect both static and dynamic occlusion accurately. On the other hand, the frame prediction model employs a time-series generator based on LSTM to effectively reconstruct the occluded frames. Both the models have been made available online for further comparison and analysis. To the best of knowledge, ours is the first work that achieves a significantly high Rank 1 accuracy even on the challenging TUM-IITKGP data set with both static and dynamic occlusion. In future, it may be studied if a bi-directional LSTM model can help in obtaining better quality of reconstructed frames. Also, our work can be extended to carry out recognition in the presence of co-variate conditions such as carrying bag, wearing coat, etc., as well as perform open-set recognition.
%Use of deep learning has beaten the state of the art by attaining pretty high accuracy score.

\section*{Acknowledgements} The authors would like to thank NVIDIA for supporting their research with a Titan Xp GPU.

\bibliographystyle{ieeetr}
\bibliography{gaitbib.bib}
\end{document}